\documentclass{article}

\usepackage{microtype}
\usepackage{graphicx}
\usepackage{subfigure}
\usepackage{booktabs} 

\usepackage[hyphens]{url}
\usepackage{hyperref}
\usepackage{multirow}

\usepackage[accepted]{icml2023}

\usepackage{amsmath}
\usepackage{amssymb}
\usepackage{mathtools}
\usepackage{amsthm}

\usepackage[capitalize,noabbrev]{cleveref}

\theoremstyle{plain}

\theoremstyle{definition}

\theoremstyle{remark}

\usepackage[textsize=tiny]{todonotes}

\icmltitlerunning{In-Context Learning Functions with Varying Number of Minima}

\begin{document}

\twocolumn[
\icmltitle{In-Context Learning Functions with Varying Number of Minima}



\icmlsetsymbol{equal}{*}

\begin{icmlauthorlist}
\icmlauthor{David Oniani}{pitt}
\icmlauthor{Yanshan Wang}{pitt}
\end{icmlauthorlist}

\icmlaffiliation{pitt}{University of Pittsburgh, Pittsburgh, PA}

\icmlcorrespondingauthor{Yanshan Wang}{yanshan.wang@pitt.edu}

\icmlkeywords{Artificial Intelligence, Large Language Models, In-Context Learning, Functions}

\vskip 0.3in
]



\printAffiliationsAndNotice{}  


\DeclarePairedDelimiter\abs{\lvert}{\rvert}
\DeclarePairedDelimiter\ceil{\lceil}{\rceil}
\DeclarePairedDelimiter\floor{\lfloor}{\rfloor}
\DeclarePairedDelimiter\norm{\lVert}{\rVert}
\newcommand{\reals}{\mathbb{R}}
\newcommand{\E}{\mathbb{E}}
\newcommand{\F}{\mathcal{F}}
\newcommand{\X}{\mathcal{X}}
\renewcommand{\L}{\mathcal{L}}


\begin{abstract}
  Large Language Models (LLMs) have proven effective at In-Context Learning (ICL), an ability that
  allows them to create predictors from labeled examples. Few studies have explored the interplay
  between ICL and specific properties of functions it attempts to approximate. In our study, we use
  a formal framework to explore ICL and propose a new task of approximating functions with varying
  number of minima. We implement a method that allows for producing functions with given inputs as
  minima. We find that increasing the number of minima degrades ICL performance. At the same time,
  our evaluation shows that ICL outperforms 2-layer Neural Network (2NN) model. Furthermore, ICL
  learns faster than 2NN in all settings. We validate the findings through a set of few-shot
  experiments across various hyperparameter configurations.
\end{abstract}

\section{Introduction}

The topic of Large Language Models (LLMs) has become a blooming area of research, with studies
ranging from cybersecurity to finance~\cite{gupta2023,xiang2023} and from the legal industry to
healthcare~\cite{zhang2023,clusmann2023}. As a result, significant efforts have been made to improve
LLMs, including research in latency~\cite{wang2023tabi}, ethics~\cite{oniani2023military}, and
evaluation~\cite{liang2023holistic}. This has also led to the development of various techniques
specific to LLMs, such as prompting and In-Context Learning (ICL).

\begin{figure*}[ht]
  \vskip 0.2in
  \begin{center}
    \centerline{\includegraphics[width=\linewidth]{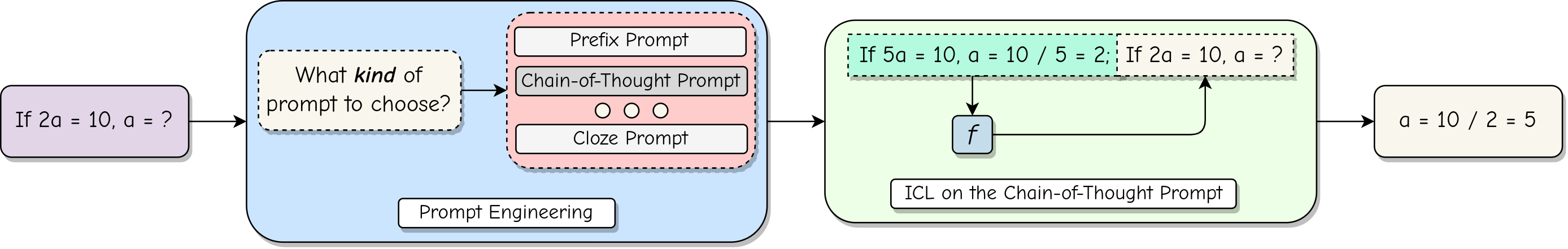}}
    \caption{LLM Pipeline: From a Query to the Output. In the Prompt Engineering step, the proper
      prompting approach is determined. While performing In-Context Learning (ICL), a function is
      constructed and then used to approximate the answer to the given query.}
    \label{prompting_and_icl}
  \end{center}
  \vskip -0.2in
\end{figure*}

Prompting is a novel paradigm, where textual prompts are used to model downstream tasks as those
typically solved during pre-training~\cite{liu2023}. The emergence of this paradigm gave rise to the
subfield of prompt engineering~\cite{reynolds2021}, which aims to find aligning prompts. As a
result, many prompt engineering approaches have been developed, including Input-Output
(IO)~\cite{ouyang2022}, Chain-of-Thought~(CoT)~\cite{wei2022}, Tree-of-Thought
(ToT)~\cite{yao2023tree}, Models-Vote~(MVP)~\cite{oniani2023large}, and
Self-Consistency~(SC)~\cite{wang2023} prompting.  Since generative LLMs make predictions based on
the given prompt (i.e., the context), there is a natural relationship between prompt engineering and
ICL (illustrated in Figure~\ref{prompting_and_icl}). The IO prompting paper introduced InstructGPT,
a model that was trained to follow instructions, which was one of the first works to popularize ICL.
The term was originally introduced in the GPT-3~paper~\cite{brown2020}.

While sometimes described as a form of prompt engineering, ICL is a learning method that relies on a
given context to learn and make predictions. Interestingly, ICL is a post-training approach and does
not require parameter updates, which is one of the primary reasons it is the most popular strategy
for interacting with LLMs. Recent work has shown that ICL can also be viewed as a higher-order
function that can construct inputs from labeled, input-output
examples~\cite{garg2022what,akyürek2023what}. Figure~\ref{prompting_and_icl} also illustrates this
idea.

While the advances in LLMs gave rise to ICL research, the current understanding of ICL and its
capabilities is limited. Furthermore, few studies have explored the relationship between ICL and
specific properties of the function it aims to approximate. In this paper, we would like to propose
a new ICL task: approximating functions with varying number of minima. To do this, we implement a
function that turns its inputs into another function with the given inputs as minima. Furthermore,
we compare the performance of ICL to that of a 2-layer Neural Network (2NN).

\subsection{Contributions}

\begin{enumerate}
  \item We propose a novel ICL task of approximating functions with varying number of minima. Our
        experiments show that increasing the number of minima degrades ICL performance. Despite
        this, ICL outperformed 2NN across a variety of configurations.
  \item We implement a general approach for producing functions with the given minima.
  \item Finally, we release the codebase\footnote{\url{https://github.com/PittNAIL/icl-minima}}
        that fully reproduces the results, and that can be run in an intuitive manner.
\end{enumerate}

\section{Related Work}

\subsection{Prompt Engineering}

The success and popularity of LLMs facilitated both theoretical and experimental studies on prompt
engineering. IO~prompting~\cite{ouyang2022} is one of the simplest prompt engineering methods, where
one or more examples are given to the language model before the final query.
CoT~prompting~\cite{wei2022} uses a series of reasoning steps to guide the LLM to the proper output.
ToT~\cite{yao2023tree} builds a tree out of prompts that can later be traversed using algorithms,
such as Breadth-First Search (BFS) and Depth-First Search (DFS). SC~\cite{wang2023} and
MVP~\cite{oniani2023large} are both ensemble prompting approaches. SC prompts the same model
multiple times, while MVP prompts multiple models once. Automatic prompt engineering approaches,
such as AutoPrompt~\cite{shin-etal-2020-autoprompt} and PromptGen~\cite{zhang-etal-2022-promptgen},
have also been proposed. However, some studies found that automated prompting does not consistently
outperform manual prompting~\cite{zhou-etal-2023-revisiting}.

\subsection{In-Context Learning}

While ICL is a novel area of research, a number of studies have already been conducted on the topic.
\citet{garg2022what} investigated ICL by approximating simple function classes. Their study showed
that the data and input dimensionality can significantly impact ICL performance. Additionally, they
found that curriculum learning improves training and leads to faster convergence.
\citet{akyürek2023what} considered linear functions in their study of ICL. They proved that
transformers can implement learning algorithms for linear models based on gradient descent and
closed-form ridge regression. \citet{li2023} explored the interplay between compositionality and
ICL. Their experiments demonstrated that storing intermediate Multilayer Perceptron (MLP) layer
outputs improves ICL performance. \citet{raventós2023pretraining} examined ICL performance on linear
regression while varying the diversity of tasks in the pre-training dataset.
\citet{bhattamishra2023understanding} studied ICL by approximating discrete functions, such as
conjunction, disjunction, and parity.

\begin{figure*}[ht]
  \vskip 0.2in
  \begin{center}
    \centerline{\includegraphics[width=\linewidth]{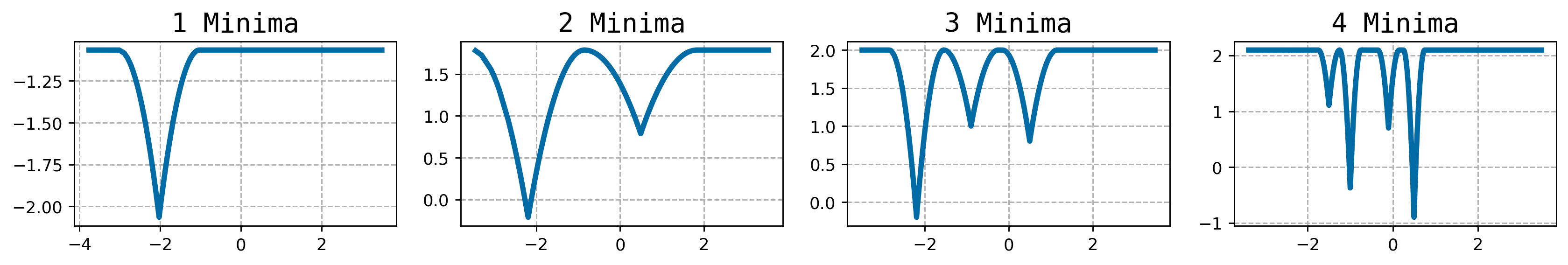}}
    \caption{Example generated functions, with one, two, three, and four minima. Plots represent the
      function described in Equation \ref{eq:6}. In the examples above, we have \(c = 1\) and the
      differentiability parameter \(\beta = 2\), which is consistent with the values used in the
      experiments.}
    \label{example_functions_with_minima}
  \end{center}
  \vskip -0.2in
\end{figure*}

\section{Preliminaries}

\subsection{Formal Framework for In-Context Learning}

In this paper, we use the formal framework introduced by \citet{garg2022what}. Utilizing the
framework, we develop the theoretical and experimental setup for the study.

A model can in-context learn a function class \(\F\) if it can approximate ``most'' of the functions
\(f \in \F\). Let \(D_\F\) and \(D_\X\) be distributions over functions in \(\F\) and inputs,
respectively. We can then define a prompt \(P\) as follows:

\begin{equation}
  P = \underbrace{x_1, f(x_1), \dots, x_n, f(x_n)}_{\text{input-output instruction sequence}}, x_{\text{query}}
\end{equation}

s.t. \(f\) is from \(\F\) and \(x_1, \dots, x_n,  x_{\text{query}}\) are i.i.d. from \(D_\X\).

Suppose \(M\) is an LLM, and let \(\L\) be the loss function appropriate for a task. Then we say
that the \(M\) can in-context learn function class \(\F\) up to \(\varepsilon\) if the following
holds:

\begin{equation}
  \E_P[{\L(M(P), f(x_{\text{query}}))}] \leq \varepsilon
\end{equation}

With this framework, we are able to define and evaluate concrete functions and function classes.

\subsection{Producing Functions with Varying Number of Minima}

Simple functions can be combined to obtain a more complex function. This is the method we will
employ for building a parametrized function with the given minima. More specifically, we will create
such a function by a linear combination of multiple similar functions. The approach was originally
inspired by the Cross Validated post~\cite{445976}. We then extended it, generalized the supported
interval, and implemented it in Python\footnote{\url{https://www.python.org/}}.

Suppose we have a set of \(n\) minima \(Y_\text{minima} = \{y_1^\prime, \dots, y_n^\prime\}\) and
the corresponding set of points where these minima are achieved \(X_\text{minima} = \{x_1^\prime,
\dots, x_n^\prime\}\). Consider a piecewise function similar to the Probability Density Function (PDF)
for the beta distribution when its parameter \(\alpha = 1\)~\cite{johnson1995-oe}. Initially, we will
support it for the arbitrary closed interval \([-c, c]\) s.t. \(c \in \reals\):

\begin{equation}
  g(x, c, \beta) = \begin{dcases}
                     (c - x)^{\beta} & \text{if } x \in [-c, c], \beta \geq 1\\
                     0               & \text{otherwise}
                   \end{dcases}
\end{equation}

This function has special properties, which provide several benefits. First, similar to the
PDF~\cite{nistsematechhandbook}, the function can be easily extended to support an arbitrary
interval. Second, we can control its differentiability via \(\beta\) parameter. Third,
it can be implemented efficiently. Our vectorized PyTorch\footnote{\url{https://pytorch.org/}}
implementation of the function has shown to be highly performant (see Appendix~\ref{sec:vecfun}).
Finally, it can be easily visualized, which will help reason about the tasks.

As already mentioned, we can support an arbitrary interval. This can be done by specifying the
location and scale parameters \(a\) and \(b\), respectively. Note that this is a general approach
that will work for any PDF or similar functions. We can define a function \(h\) that can support a
closed interval of the form \([a - b \times c, a + b \times c]\) as follows:

\begin{equation}
  h(x, c, \beta, a, b) = g(x_{\text{scaled}}) = g\Bigg(\frac{\abs{x - a}}{b}\Bigg)
\end{equation}

The transformation is similar to that performed in the PDF of the Laplace
distribution~\cite{geraci2018}. It reflects the graph over the y-axis, resulting in a function that
looks like the PDF of the Laplace distribution.

Notice that \(h\) achieves a unique maximum at \(x = a\) and its differentiability is
still controlled by \(\beta\). Now, given the set of minima \(Y_\text{minima}\) with the
corresponding points \(X_\text{minima}\), the following linear combination of functions will satisfy
the requirements:

\begin{equation}
  h^{\prime}(x) = \sum_{i = 1}^n y_i^\prime \times h(x, c, \beta, x_i^\prime, b)\label{eq:5}
\end{equation}

where \(b\) is half of the minimum distance among all distances between points in \(X_\text{minima}\):

\begin{equation}
  b = \frac{\min{\{\text{dist}(x_i, x_j) \mid i, j \in X_\text{minima}, i \neq j\}}}{2}
\end{equation}

Now, we can also ensure that \(Y_\text{minima}\) are the \textit{local} minima. In order to do
this, we will first need to introduce a parameter that is greater than any minima:

\begin{equation}
  R = \max(Y_{\text{minima}}) + C
\end{equation}

Parameter \(C\) is a constant that can be ``tuned,'' but for our purposes, we can use \(C = 1\).
This ensures that \(R\) is greater than any point in \(Y_{\text{minima}}\). We can then use \(R\)
to add a negative multiplicative factor to every function in the sum. Finally, we obtain the
following function:

\begin{equation}
  f(x) = R + \sum_{i = 1}^n (y_i^\prime - R) \times h(x, c, d, x_i^\prime, b)\label{eq:6}
\end{equation}

Such a construction will guarantee that the minima will be achieved at the given points.
Furthermore, the minima points will necessarily be the \textit{local} minima. And there is only one,
global maximum achieved at \(R\). In this study, we set \(c = 1\) and \(\beta = 2\).
Figure~\ref{example_functions_with_minima} shows plots of such functions with various numbers of
minima.

It should be noted that there are other ways to construct functions with the arbitrary number of
minima. Implementing such functions and evaluating ICL performance could also be a potential future
direction of research.


\section{Experiments}

\subsection{Models}

In our experiments, we used four GPT-2-based~\cite{radford2019language} models from
Hugging~Face\footnote{\url{https://huggingface.co/docs/transformers/model_doc/gpt2}}.
Table~\ref{models} lists the models. Note that the number of parameters was affected by
Safetensors\footnote{\url{https://huggingface.co/docs/safetensors/index}}. As a result, parameter
counts for the models are higher than they would have been without Safetensors.

\begin{table}[ht]
  \caption{GPT-2-based transformer models used in the experiments. We describe them in terms of
    number of embedding dimensions, number of heads, number of layers, and the number of
    parameters.}
  \label{models}
  \vskip 0.15in
  \begin{center}
    \begin{small}
      \begin{sc}
        \begin{tabular}{lcccc}
          \toprule
          Model    & \#Embed & \#Heads & \#Layers & \#Params\\
          \midrule
          Pico     & 32      & 1       & 1        & \(1.7\text{\textnormal e+}5\)\\ 
          Tiny     & 64      & 2       & 3        & \(3.4\text{\textnormal e+}5\)\\ 
          Small    & 128     & 4       & 6        & \(7.8\text{\textnormal e+}5\)\\ 
          Standard & 256     & 8       & 12       & \(22.6\text{\textnormal e+}5\)\\ 
          \bottomrule
        \end{tabular}
      \end{sc}
    \end{small}
  \end{center}
  \vskip -0.1in
\end{table}

The maximum context size for all of the models was set to 1,024. The context was fully utilized
in the experiments.

For the baseline model, we chose a 2-layer Neural Network (2NN) with 100 intermediate features and
ReLU~\cite{fukushima1975} activation function.

\subsection{Training and Evaluation}

We used Mean Squared Error Loss (MSE) as the loss function in all of our experiments. As for the
optimizer, we used AdamW~\cite{kingma2017adam,sashank2018} with the learning rate of
\(1\text{\textnormal e-}4\).

\begin{table}[ht]
  \caption{Hyperparameters for the few-shot experiments.}
  \label{hyperparameters}
  \vskip 0.15in
  \begin{center}
    \begin{small}
      \begin{sc}
        \begin{tabular}{ll}
          \toprule
          Hyperparameter & Options\\
          \midrule
          Shots          & 8, 16, 32\\
          Minima         & 1, 2, 3, 4\\
          Epochs         & 1,000\\
          Learning Rate  & \(1\text{\textnormal e-}4\)\\
          \bottomrule
        \end{tabular}
      \end{sc}
    \end{small}
  \end{center}
  \vskip -0.1in
\end{table}

We varied the number of minima in the generated functions and evaluated functions with 1, 2, 3, and
4 minima. The number of shots was also varied and included 8, 16, and 32 shots. Inputs were sampled
from the Gaussian distribution \(\mathcal{N}(0, 1)\). Every LLM model was trained and evaluated on
the datasets of equal size. In all cases, the training was done for 1,000 epochs.
Table~\ref{hyperparameters} lists the hyperparameters.

The hyperparameters for 2NN were the same as those for the transformer-based models and are shown in
Table~\ref{hyperparameters}. Training and evaluation datasets were also the same. For more
information about the training and evaluation setup, see
Appendix~\ref{sec:training_evaluation_setup}.

All trainable layers for the models were reset before training and hence, none of the GPT-2 based
models had any pre-training or warm-starting advantage.

The training jobs were run on a single Quadro RTX 8000 GPU. The experimental results presented in
the study can be reproduced on the same or similar GPU in approximately 3 hours.


\section{Results}

We divide the experimental results into two subsections: training and evaluation. In the training
subsection, we analyze loss plots, discuss training behavior, and compare MSE values. In the
evaluation subsection, we directly compare MSE values on the evaluation data. Note that we sampled
both training and evaluation data from the Gaussian \(\mathcal{N}(0, 1)\) distribution.

\subsection{Training}
\label{subsec:training}

In this subsection, we analyze the results across model types, prompt\footnote{Terms ``prompt'' and
``shot'' are used interchangeably.} counts, and the number of minima.

\subsubsection{Pico}

For Pico model, increasing the number of shots decreased the loss and improved the convergence. The
behavior was similar for 2NN. Increasing the number of minima had marginal effects on the
convergence rate or the final MSE values. In all experiments, the MSE loss for Pico decreased
significantly faster than that of 2NN, leading to lower overall MSE.

See Figure~\ref{loss:pico} for the loss plots of the Pico model with inputs sampled from the
Gaussian distribution \(\mathcal{N}(0, 1)\).

\subsubsection{Tiny}

The rate of convergence for Tiny was even faster than that for Pico. As a result, MSE values for Tiny were even smaller than those of Pico. In all experiments, the loss for Tiny decreased faster than the MSE for 2NN. The number of minima did not have a significant impact on convergence. Increasing the number of epochs led to a decrease in the overall MSE value.

See Figure~\ref{loss:tiny} for the loss plots of the Tiny model with inputs sampled from the
Gaussian distribution \(\mathcal{N}(0, 1)\).

\subsubsection{Small}

Similar to smaller models, loss for Small decreased more quickly than for 2NN, resulting in MSE
values even smaller than those of both Pico and Tiny. In general, increasing the number of prompts
decreased MSE. The convergence rate and the overall MSE were similar, irrespective of the number of
minima.

See Figure~\ref{loss:small} for the loss plots of the Small model with inputs sampled from the
Gaussian distribution \(\mathcal{N}(0, 1)\).

\subsubsection{Standard}

Standard exhibited the same behavior as the previous models, with a faster convergence rate and
smaller loss values. Changing the number of minima did not result in drastic changes in the
convergence rate or the overall MSE. Loss decreased as the number of prompts increased.

See Figure~\ref{loss:standard} for the loss plots of the Standard model with inputs sampled from the
Gaussian distribution \(\mathcal{N}(0, 1)\).

In summary, ICL has shown to be robust against increases in the number of minima. This has been
three prompt sizes and four minima. It has also shown remarkable convergence across the experiments.

\subsection{Evaluation}

Similar to Subsection~\ref{subsec:training}, we will discuss the results considering model type, the
number of prompts, and the number of minima.

\subsubsection{Pico}

In the case of Pico, as the number of prompts increased, the evaluation loss decreased. The only
exception was when the number of minima was 3. The behavior was not the same for 2NN, but the model
did not exhibit performance degradation. Increasing the number of minima increased the MSE and
decreased the overall performance.

For results, see Table~\ref{tab:pico}.

\subsubsection{Tiny}

As for Tiny, the behavior was similar to Pico but with smaller loss values. Increasing the number of
prompts decreased the loss. ICL performance degraded as the number of minima increased. Overall,
Tiny outperformed 2NN, with the exception being the case when the number of minima is 4.

For results, see Table~\ref{tab:tiny}.

\subsubsection{Small}

Small showed even smaller loss values and better overall performance. The behavior across epochs and
the number of minima was similar to those of Pico and Tiny. Small model consistently outperformed
2NN.

For results, see Table~\ref{tab:small}.

\subsubsection{Standard}

Standard had the best overall performance. Increasing the number of prompts decreased the MSE. In
general, increasing the number of minima resulted in smaller MSE values. Similar to Small, Standard
outperformed 2NN across all benchmarks.

For results, see Table~\ref{tab:standard}.

Overall, the performance for approximating a function with one minima was the best. As the number of
minima increased, the MSE values increased, and the performance degraded. Increasing the number of
prompts decreased the MSE.

Smaller training loss may be due to faster convergence. In other words, since the loss for the model
decreased quickly in the beginning, the overall MSE was smaller.


\section{Limitations and Future Work}

Our work has several limitations that could lead to the potential future work. First, in our study
of ICL, we constructed a function in a specific manner. Implementing other approaches for building
functions with an arbitrary number of minima and evaluating ICL could be an interesting future
research direction. Second, other choices for hyperparameters (e.g., more minima, fewer epochs,
different choice of the \(\beta\) parameter, etc.) could be another avenue for exploration. Third,
constructing out-of-distribution prompts for evaluation can provide further details on the
robustness of ICL. Fourth, given sufficient resources, it could be worthwhile to consider larger
(e.g., GPT-3.5 or GPT-4-scale) transformer-based LLMs. Fifth, incorporating prompting approaches
beyond IO (e.g., SC) may show promising results. Finally, an ability to generate functions with the
given minima also allows for defining new tasks (e.g., counting minima, approximating minima
coordinates, etc).


\section{Conclusion}

In this paper, we proposed a new ICL task of approximating functions with varying number of minima.
Our research showed that, in general, increasing the number of minima degrades ICL performance. At
the same time, ICL outperformed 2NN model. We validated the findings through a series of experiments
with different hyperparameter configurations and analyzed both the training and evaluation results.
In addition, we implemented a performant, vectorized function for producing functions with the given
minima. Finally, we also released the codebase that fully reproduces the results.


\section{Ethics and Societal Impact Statement}

This paper presents work whose goal is to advance the field of Machine Learning. There are many
potential societal consequences of our work, none which we feel must be specifically highlighted
here.


\bibliography{main}
\bibliographystyle{icml2023}


\newpage
\appendix
\onecolumn

\counterwithin*{equation}{section}
\renewcommand\theequation{\thesection\arabic{equation}}

\renewcommand{\thefigure}{A.\arabic{figure}}
\setcounter{figure}{0}

\renewcommand{\thetable}{A.\arabic{table}}
\setcounter{table}{0}

\section{Appendix}
\label{sec:appendix}

\subsection{Pseudocode for the Vectorized Function Implementation}
\label{sec:vecfun}

Algorithm~\ref{alg:helper} implements a helper function that will later be used in
Algorithm~\ref{alg:genmin}. The helper function takes a single input and applies the differentiable
function discussed previously.

\begin{algorithm}[H]
  \caption{A helper function that will be used for creating a linear combination of functions in the generator}
  \label{alg:helper}
  \begin{algorithmic}
    \STATE \textbf{Input:} data \(x\), initial support interval parameter \(c\), differentiability parameter \(\beta\), location parameter \(a\), scale parameter \(b\)
    \STATE \textbf{Output:} function evaluated at \(x\) with the default values of \(c = 1.0\), \(\beta = 2.0\), \(a = 0.0\) and \(b = 1.0\)
    \STATE
    \FUNCTION{h\((x, c, \beta, a, b)\)}
      \STATE \textbf{return} \((c - min(abs((x - a) / b), c))^\beta\)
    \ENDFUNCTION
  \end{algorithmic}
\end{algorithm}

Algorithm~\ref{alg:genmin} makes use of the helper, and produces the function that is supported at
the given points and minima. Note that the points and minima can be sampled from some arbitrary
distributions.

\begin{algorithm}[H]
  \caption{Broadcasted function generator that produces functions with the given minima}
  \label{alg:genmin}
  \begin{algorithmic}
    \STATE \textbf{Input:} second-order tensor \(m\) with pairs of minima points, input points \(p\)
    \STATE \textbf{Output:} function supported on points \(ps\) s.t. minima is at \(m\)
    \STATE \textbf{Preliminaries:} Function l2dist computes a vector of L2 distances between all pairs different points
    \STATE \textbf{Assumptions:} Broadcasting
    \STATE
    \FUNCTION{generator\((m, p)\)}
      \STATE \(x,\) \(y = \text{transpose}(m)\)
      \STATE \(b = \text{min}(\text{l2dist}(x)) / 2\)
      \STATE \(R = \text{max}(y) + 1\)
      \STATE \(out = \textbf{h}(\text{repeat}(\text{unsqueeze}(p, \text{dim}=1), \text{dims}=(1, len(x))), a=x, b=b)\)
      \STATE \textbf{return} \(R + \text{sum}((y - R) * out, \text{dim}=1)\)
  \ENDFUNCTION
  \end{algorithmic}
\end{algorithm}

The first step of the algorithm is to get \(x\) and \(y\) coordinates. This can be done by
transposing the input matrix and unpacking the points. The next step is computing the scale, which
is done by computing euclidean (L2) distances, finding the minimum distance, and dividing it by 2
(i.e., finding the midpoint of the minimum).

The pseudocode was implemented in Python using PyTorch.

\subsection{Training and Evaluation Setup}
\label{sec:training_evaluation_setup}

In both training and evaluation, we have one-dimensional inputs and outputs. Since a single prompt
(i.e., a shot) contains 1,024 tokens, this results in 512 input-output pairs. Therefore, 8, 16, and
32-shot experiments have 4,096, 8,192, and 16,384 input-output pairs, respectively. In the case of
the transformer, they are assembled in the prompt \(x_1, y1, \dots, x_n, y_n\). For 2NN, \(x1,
\dots, x_n\) are used as features and \(y_1, \dots, y_n\) as labels.

Both transformer-based LLMs and 2NN models were trained via gradient descent. Evaluation was
separate from training and had the same number of samples.

For LLMs, the dropout was fully disabled. Specifically, parameters \texttt{attn\_pdrop},
\texttt{embd\_pdrop}, and \texttt{use\_cache} were set to
0\footnote{\url{https://huggingface.co/docs/transformers/model_doc/gpt2}}.

\subsection{Experimental Results}
\label{sec:experimental_results}

\begin{figure}[H]
  \begin{center}
    \centerline{\includegraphics[width=\linewidth]{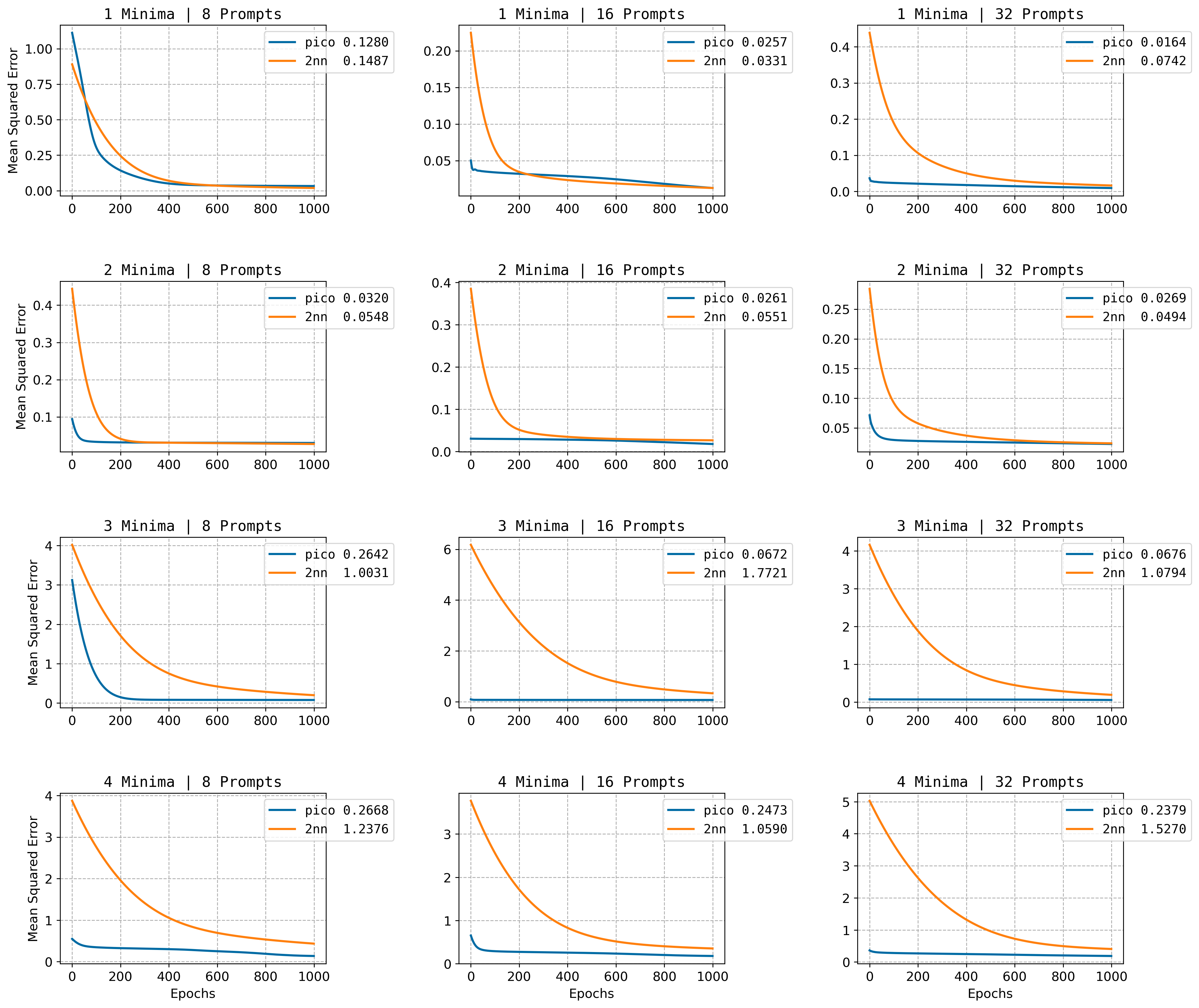}}
    \caption{Mean Squared Error (MSE) over epochs for Pico model with inputs sampled from the Gaussian distribution \(\mathcal{N}(0, 1)\). Note that the MSE values for Pico and 2NN shown in the legend are computed across all epochs (i.e., averaged out across 1,000 epochs).}
    \label{loss:pico}
  \end{center}
\end{figure}

\clearpage

\begin{figure*}
  \vspace*{\fill}
  \begin{center}
    \centerline{\includegraphics[width=\linewidth]{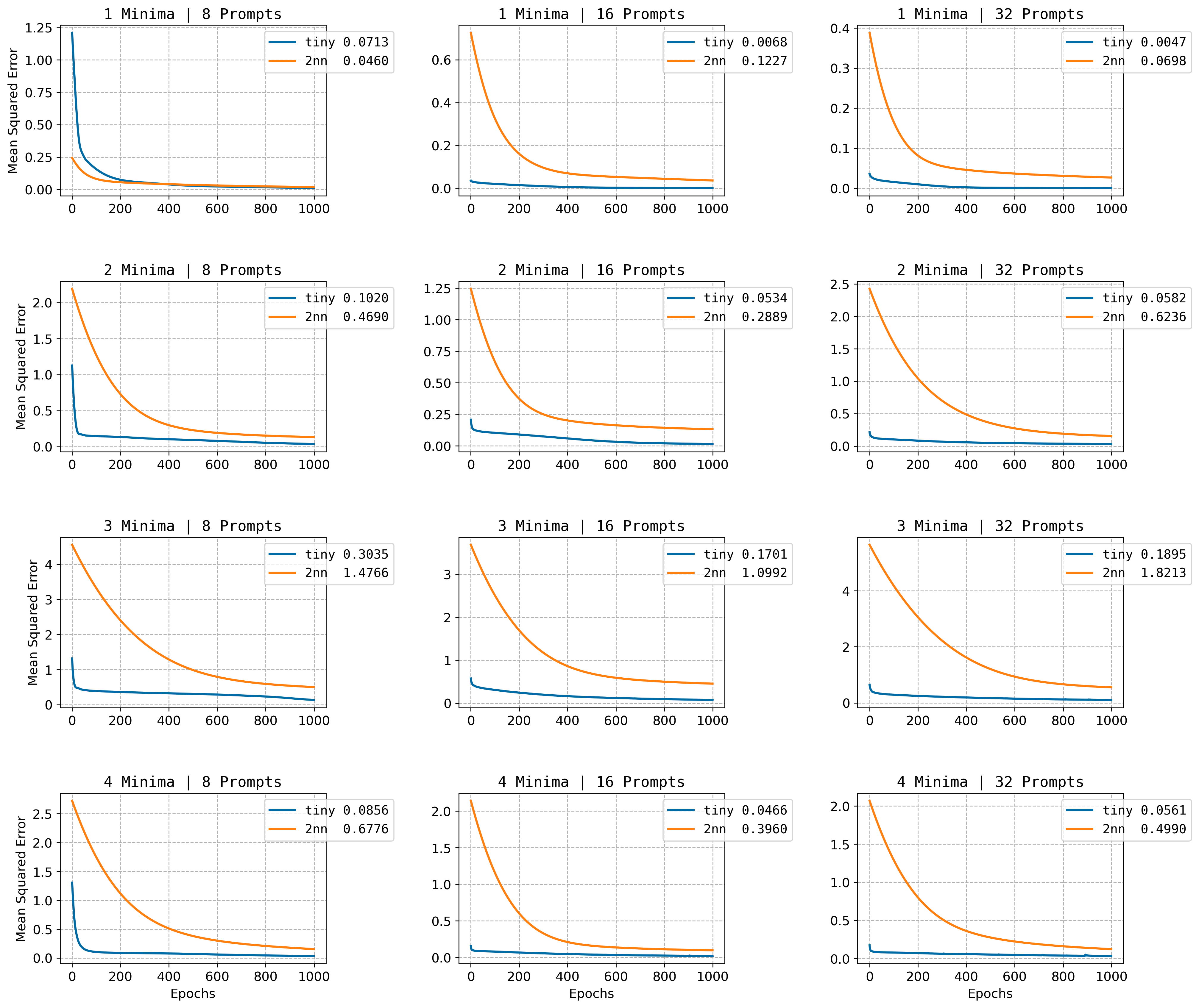}}
    \caption{Mean Squared Error (MSE) over epochs for Tiny model with inputs sampled from the Gaussian distribution \(\mathcal{N}(0, 1)\). Note that the MSE values for Tiny and 2NN shown in the legend are computed across all epochs (i.e., averaged out across 1,000 epochs).}
    \label{loss:tiny}
  \end{center}
  \vspace*{\fill}
\end{figure*}

\clearpage

\begin{figure*}
  \vspace*{\fill}
  \begin{center}
    \centerline{\includegraphics[width=\linewidth]{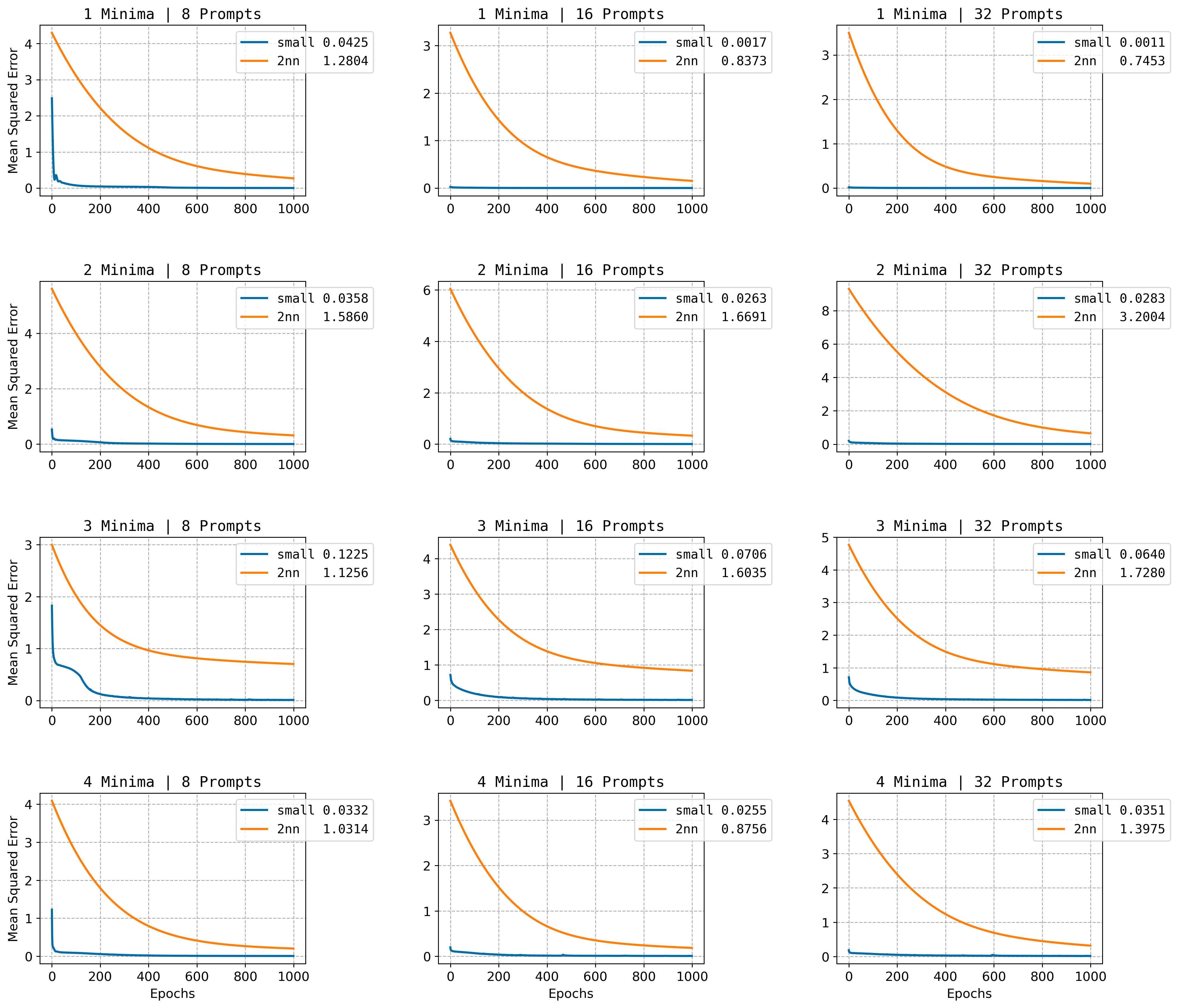}}
    \caption{Mean Squared Error (MSE) over epochs for Small model with inputs sampled from the Gaussian distribution \(\mathcal{N}(0, 1)\). Note that the MSE values for Small and 2NN shown in the legend are computed across all epochs (i.e., averaged out across 1,000 epochs).}
    \label{loss:small}
  \end{center}
  \vspace*{\fill}
\end{figure*}

\clearpage

\begin{figure*}
  \vspace*{\fill}
  \begin{center}
    \centerline{\includegraphics[width=\linewidth]{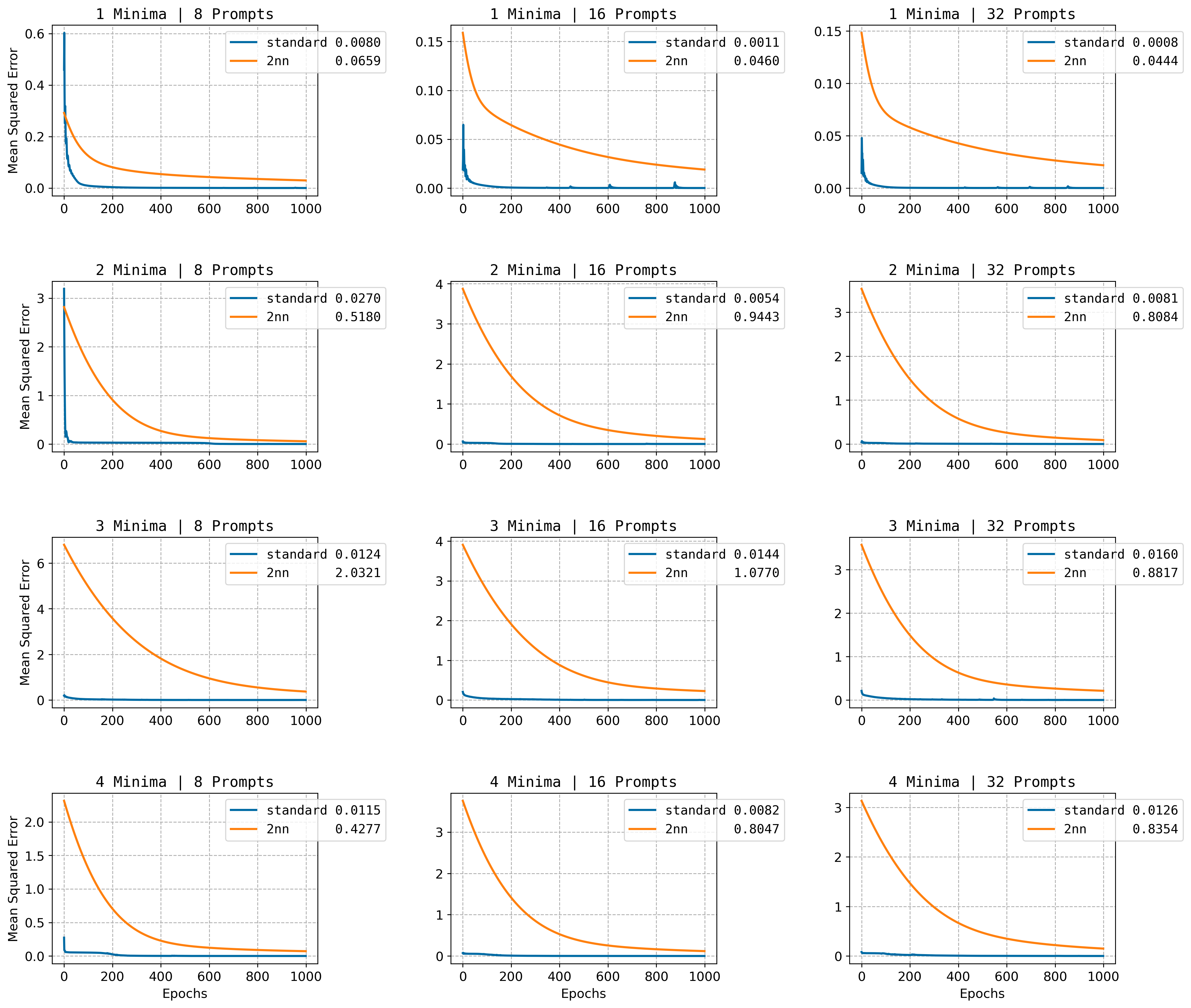}}
    \caption{Mean Squared Error (MSE) over epochs for Standard model with inputs sampled from the Gaussian distribution \(\mathcal{N}(0, 1)\). Note that the MSE values for Standard and 2NN shown in the legend are computed across all epochs (i.e., averaged out across 1,000 epochs).}
    \label{loss:standard}
  \end{center}
  \vspace*{\fill}
\end{figure*}

\clearpage

\begin{table}[H]
  \caption{Mean Squared Error (MSE) of the Pico model with inputs sampled from the Gaussian distribution \(\mathcal{N}(0, 1)\).}
  \label{tab:pico}
  \vskip 0.15in
  \begin{center}
    \begin{small}
      \begin{sc}
        \resizebox{\linewidth}{!}{
          \begin{tabular}{l|ccc|ccc|ccc|ccc}
            \toprule
            \multicolumn{1}{c}{Model} & \multicolumn{3}{c}{Minima \(= 1\)} & \multicolumn{3}{c}{Minima \(= 2\)} & \multicolumn{3}{c}{Minima \(= 3\)} & \multicolumn{3}{c}{Minima \(= 4\)}\\
            \midrule
             & 8 & 16 & 32  & 8 & 16 & 32  & 8 & 16 & 32  & 8 & 16 & 32 \\
            \midrule
            2NN      & 0.0195 & 0.0132 & 0.0163 & 0.0243 & 0.0248 & 0.0238 & 0.1846 & 0.3254 & 0.1897 & 0.4515 & 0.3594 & 0.4124\\
            Pico     & 0.0354 & 0.0158 & 0.0099 & 0.0278 & 0.0249 & 0.0240 & 0.0677 & 0.0652 & 0.0911 & 0.4463 & 0.2648 & 0.2476\\
            \bottomrule
          \end{tabular}
        }
      \end{sc}
    \end{small}
  \end{center}
  \vskip -0.1in
\end{table}

\begin{table}[H]
  \caption{Mean Squared Error (MSE) of the Tiny model with inputs sampled from the Gaussian distribution \(\mathcal{N}(0, 1)\).}
  \label{tab:tiny}
  \vskip 0.15in
  \begin{center}
    \begin{small}
      \begin{sc}
        \resizebox{\linewidth}{!}{
          \begin{tabular}{l|ccc|ccc|ccc|ccc}
            \toprule
            \multicolumn{1}{c}{Model} & \multicolumn{3}{c}{Minima \(= 1\)} & \multicolumn{3}{c}{Minima \(= 2\)} & \multicolumn{3}{c}{Minima \(= 3\)} & \multicolumn{3}{c}{Minima \(= 4\)}\\
            \midrule
             & 8 & 16 & 32  & 8 & 16 & 32  & 8 & 16 & 32  & 8 & 16 & 32 \\
            \midrule
            2NN      & 0.0176 & 0.0355 & 0.0258 & 0.1315 & 0.1276 & 0.1464 & 0.5017 & 0.4415 & 0.5441 & 0.1685 & 0.1060 & 0.1292\\
            Tiny     & 0.0163 & 0.0015 & 0.0006 & 0.1525 & 0.1023 & 0.0766 & 0.4448 & 0.4192 & 0.2484 & 0.1578 & 0.1557 & 0.1373\\
            \bottomrule
          \end{tabular}
        }
      \end{sc}
    \end{small}
  \end{center}
  \vskip -0.1in
\end{table}

\begin{table}[H]
  \caption{Mean Squared Error (MSE) of the Small model with inputs sampled from the Gaussian distribution \(\mathcal{N}(0, 1)\).}
  \label{tab:small}
  \vskip 0.15in
  \begin{center}
    \begin{small}
      \begin{sc}
        \resizebox{\linewidth}{!}{
          \begin{tabular}{l|ccc|ccc|ccc|ccc}
            \toprule
            \multicolumn{1}{c}{Model} & \multicolumn{3}{c}{Minima \(= 1\)} & \multicolumn{3}{c}{Minima \(= 2\)} & \multicolumn{3}{c}{Minima \(= 3\)} & \multicolumn{3}{c}{Minima \(= 4\)}\\
            \midrule
             & 8 & 16 & 32  & 8 & 16 & 32  & 8 & 16 & 32  & 8 & 16 & 32 \\
            \midrule
            2NN      & 0.2654 & 0.1513 & 0.0986 & 0.3176 & 0.3388 & 0.6532 & 0.6747 & 0.7911 & 0.8528 & 0.2027 & 0.1764 & 0.3088\\
            Small    & 0.0069 & 0.0008 & 0.0002 & 0.1599 & 0.1344 & 0.1011 & 0.3607 & 0.2264 & 0.0761 & 0.1477 & 0.1147 & 0.1103\\
            \bottomrule
          \end{tabular}
        }
      \end{sc}
    \end{small}
  \end{center}
  \vskip -0.1in
\end{table}

\begin{table}[H]
  \caption{Mean Squared Error (MSE) of the Standard model with inputs sampled from the Gaussian distribution \(\mathcal{N}(0, 1)\).}
  \label{tab:standard}
  \vskip 0.15in
  \begin{center}
    \begin{small}
      \begin{sc}
        \resizebox{\linewidth}{!}{
          \begin{tabular}{l|ccc|ccc|ccc|ccc}
            \toprule
            \multicolumn{1}{c}{Model} & \multicolumn{3}{c}{Minima \(= 1\)} & \multicolumn{3}{c}{Minima \(= 2\)} & \multicolumn{3}{c}{Minima \(= 3\)} & \multicolumn{3}{c}{Minima \(= 4\)}\\
            \midrule
             & 8 & 16 & 32  & 8 & 16 & 32  & 8 & 16 & 32  & 8 & 16 & 32 \\
            \midrule
            2NN      & 0.0301 & 0.0189 & 0.0218 & 0.0520 & 0.1202 & 0.0906 & 0.3710 & 0.2296 & 0.2177 & 0.0785 & 0.1142 & 0.1488\\
            Standard & 0.0038 & 0.0006 & 0.0002 & 0.0420 & 0.0391 & 0.0419 & 0.1891 & 0.1714 & 0.1266 & 0.0953 & 0.0876 & 0.0809\\
            \bottomrule
          \end{tabular}
        }
      \end{sc}
    \end{small}
  \end{center}
  \vskip -0.1in
\end{table}

\end{document}